\documentclass[letterpaper, 10 pt, conference]{ieeeconf}  

\IEEEoverridecommandlockouts                              

\usepackage{graphics} 
\usepackage{epsfig} 
\usepackage{times} 
\usepackage{amsmath} 
\usepackage{amssymb}  
\usepackage{mathtools}
\usepackage{float}
\newtheorem{remark}{Remark}{}
\newtheorem{assumption}{Assumption}{}
{}
\newtheorem{theorem}{Theorem}{}
{}
\newtheorem{problem}{Problem}{}
\usepackage{color}
\usepackage{enumerate}
\usepackage{cite}

\usepackage{tikz}
\usetikzlibrary{shapes.geometric, arrows}

\tikzstyle{block} = [draw, fill=white, rectangle, 
    minimum height=3em, minimum width=3em, text width=3.3cm, align=center]
\tikzstyle{block_small} = [draw, fill=white, rectangle, 
    minimum height=3em, minimum width=3em, text width=2cm, align=center]
\tikzstyle{block_kin} = [draw, line width=0.5mm, fill=black!10, rectangle, 
    minimum height=3em, minimum width=6em, text width=3cm, align=center]
\tikzstyle{block_nei} = [draw, dashed, line width=0.5mm, fill=blue!10, rectangle, 
    minimum height=3em, minimum width=7cm, minimum height=1cm, text width=6cm, align=center]

\tikzstyle{sum} = [draw, fill=white, circle, node distance=1cm]
\tikzstyle{point} = [coordinate]
\tikzstyle{pinstyle} = [pin edge={to-,thin,black}]

\usepackage{xcolor}

\usepackage[hidelinks]{hyperref}

\newcommand{\scalemath}[2]{\scalebox{#1}{\mbox{\ensuremath{\displaystyle #2}}}}
\newtheorem{example}{Example}

\title{\LARGE \bf
Distributed Oscillatory Guidance for Formation Flight \\of Fixed-Wing Drones
}

\author{Yang Xu*, Jes\'{u}s Bautista*, Jos\'{e} Hinojosa, H\'{e}ctor Garc\'{i}a de Marina%
\thanks{Yang Xu, Jes\'{u}s Bautista, Jos\'{e} Hinojosa, and H\'{e}ctor Garc\'{i}a de Marina are with the Department of Computer Engineering, Automation, and Robotics, and with IMAG, University of Granada, Spain. This work is supported by the ERC Starting Grant iSwarm 101076091 and the RYC2020-030090-I grant from the Spanish Ministry of Science. {\tt\small \{yang.xu, jesusbv, josehinojosa, hgdemarina\}@ugr.es}. *Yang and Jesús contributed equally.}
}

\begin{document}

\maketitle
\thispagestyle{empty}
\pagestyle{empty}

\begin{abstract}
The autonomous formation flight of fixed-wing drones is hard when the coordination requires the actuation over their speeds since they are critically bounded and aircraft are mostly designed to fly at a nominal airspeed. This paper proposes an algorithm to achieve formation flights of fixed-wing drones without requiring any actuation over their speed. In particular, we guide all the drones to travel over specific paths, e.g., parallel straight lines, and we superpose an \emph{oscillatory behavior} onto the guiding vector field that drives the drones to the paths. This oscillation enables control over the average velocity along the path, thereby facilitating inter-drone coordination. Each drone adjusts its oscillation amplitude distributively in a closed-loop manner by communicating with neighboring agents in an undirected and connected graph. A novel consensus algorithm is introduced, leveraging a non-negative, asymmetric saturation function. This unconventional saturation is justified since \emph{negative} amplitudes do not make drones travel backward or have a negative velocity along the path. Rigorous theoretical analysis of the algorithm is complemented by validation through numerical simulations and a real-world formation flight.
\end{abstract}

\section{Introduction}

Drone swarms have the tremendous potential to enhance the resilience and effectiveness of a wide range of civil and military tasks, including environmental monitoring, border control, communication infrastructure deployment, and search-and-rescue operations \cite{brambilla2013swarm,zhou2023racer,zhou2022swarm,leonard2010coordinated}. One of the greatest advantages of general robot swarms lies in their capacity for persistent operations and their ability to simplify certain logistical challenges\cite{yang2018grand,rubenstein2014programmable}. For example, they can split payloads traditionally carried by a single complex robot across multiple smaller, lower-cost, and easier-to-replace drones, thereby reducing both risk and operational costs.

Among aerial platforms, fixed-wing drones are particularly attractive due to their energy efficiency. However, controlling their speed is inherently more complex. Unlike rotorcraft, fixed-wing drones require a minimum speed to maintain lift and are typically optimized for maintaining a steady cruising speed. Adjusting speed is nontrivial, as any change affects not only forward velocity but also vertical lift, thereby impacting altitude. These dynamics introduce a tight coupling between speed and altitude, which complicates coordination. Additionally, fixed-wing platforms are subject to non-holonomic constraints, allowing only forward motion.

When fixed-wing drones operate near their nominal cruising speed and altitude is managed by a lower-level controller, typically through pitch and throttle regulation, the drone’s motion can be reasonably approximated by a 2D unicycle model with constant speed. This simplification retains the key non-holonomic constraint of forward motion while reducing the control problem to heading control only, making it well-suited for coordination and path-following tasks.
Most of the formation control algorithms in the literature for the precise coordination of unicycles assume full actuation over both heading and speed \cite{anderson2008uav,wang2007cooperative,stipanovic2004decentralized,lin2005necessary,dimarogonas2007rendezvous,marshall2006pursuit,el2012distributed,yu2022decentralized,zheng2015distributed,wang2022fixed,chen2022coordinated,chen2021coordinated}; \emph{passing the ball} to solve challenging engineering problems such as the system identification of the drone in order to have a faithful transfer to real-world flights. There are relevant distributed algorithms for the coordination of unicycles with constant speeds \cite{sepulchre2007stabilization,sepulchre2008stabilization}; however, either they do not allow precise geometric patterns or there is no evidence of real-world experiments. Regarding real-world formation flights with constant speeds, we improve our previous work in \cite{de2017circular} by allowing generic paths rather than only closed ones with a sufficiently large curvature, and achieving total control over the transitory.

In this paper, we present a distributed reactive control law to achieve formation flights of fixed-wing drones on generic paths without requiring actuation on their speeds. Distributed means that the drones only count on local information to make their decisions, and reactive means that no motion planning or optimization solvers are involved as usually employed in multi-robot coordination \cite{yu2016optimal,nageli2017real}; in fact, we will show that a simple microcontroller suffices to run our algorithm based on a novel consensus protocol. Moreover, a rigorous analysis provides the necessary mathematical guarantees to achieve the desired drone formation flight.

We begin by guiding the drones toward their assigned paths—either open or closed—, which have a parametric representation. The desired formation is achieved once the drones reach a specific relative parametric position along these paths. Assuming all drones maintain a nominal and common cruising speed—which may vary slightly in practice—, 
we regulate the progression along the path by modulating the effective path-parameter velocity. This is achieved by inducing controlled oscillations in the guiding vector field (GVF). For example, increasing the amplitude of these oscillations causes the drone to advance more slowly along the path on average, even while maintaining a nominal cruise speed.
In particular, we will command amplitudes of \textit{fixed-frequency oscillations}. The GVF to drive the drones to the desired path is based on controlling a level-set error signal \cite{kapitanyuk2017guiding,de2017guidance}, where the zero level set corresponds to the desired path. Our recent work \cite{zhou2025inverse} introduces the \textit{inverse kinematics} approach for this GVF. Specifically, this technique enables us to model the dynamics of the level-set error signal as a single integrator system; thus, facilitating precise controlled oscillations in the error signal around the desired zero level set, which we exploit to control the drones' average velocity progression along the path. Although this strategy can be applied to any generic path that admits a singularity-free inverse kinematics guiding vector field (IK-GVF) \cite{zhou2025inverse}, in this conference paper we focus on straight-line paths to simplify the presentation and highlight the core contributions.

It is important to emphasize that our approach does not enforce a constant speed on the drones; instead, it eliminates the need for speed actuation. The assumption of constant speeds across the swarm is adopted primarily for the mathematical convenience of the convergence analysis. Crucially, in real-world flight experiments, we observe that drone speeds naturally vary during transient phases, e.g., due to wind disturbances or altitude adjustments, yet the formation flight still converges robustly. This demonstrates that our algorithm inherently compensates for speed fluctuations without requiring active speed regulation.

The distributed consensus algorithm governing the oscillations employs a non-standard saturation function—\textit{shifted} along both the vertical and horizontal axes—compared to the conventional or even the more recently proposed saturation functions \cite{zou2023consensus}. Specifically, our saturation function yields zero output for zero or negative inputs and saturates at a positive maximum value for positive inputs. This output directly drives the average progress velocity along the path, which is constrained between two positive bounds, i.e., the aircraft cannot fly backwards. These bounds correspond to two scenarios; namely, zero amplitude meaning a maximum average progress velocity, and maximum amplitude meaning minimum average progress velocity. This maximum amplitude, together with the oscillation's fixed frequency, is given by the aircraft's physical limits. Note that the magnitude of the amplitude, and not its sign, determines the average velocity progress. When all drones achieve their desired inter-vehicle spacing parameters on the path, the consensus algorithm naturally drives all amplitudes to zero, stabilizing the formation flight, where the average velocity progress equals the speed of the aircraft.

This paper is organized as follows. Section \ref{sec: pre} introduces the necessary background to develop our distributed formation flight algorithm. In Section \ref{sec: results}, we design the oscillatory behavior, a consensus controller for its regulation, and a heading controller to ensure following the designated path and oscillatory behavior, thereby solving the formation flight problem for fixed-wing drones. Finally, in Section \ref{sec: exp}, we experimentally validate the proposed approach through a real-world test with two fixed-wing drones.

\section{Preliminaries and problem formulation}
\label{sec: pre}

\subsection{Notation}
Given a vector $x\in\mathbb{R}^2$, its Euclidean norm is denoted as $\|x\|$, and its unit form as $\hat x = x / \|x\|$. We define as $1_N \in\mathbb{R}^N$ the vector with only $1$ entries. In a system of $N$ robots, any quantity associated with the $i$'th robot is represented with an $i$ subindex, e.g., $x_{i,d}$ denotes the vector $x_d$ corresponding to the $i$'th robot.

\subsection{Graph Theory}

For an interaction graph $\mathcal{G} := \{\mathcal{V}, \mathcal{E}\}$, where $\mathcal{V} :=\{1,2,\cdots,N\}$ is denoted as the node set, and $\mathcal{E} \subseteq (\mathcal{V} \times \mathcal{V})$ represents the edge set. We consider only undirected graphs, i.e., the two nodes in an edge can transmit and receive information from each other. Given an ordered set of edges with cardinality $|\mathcal{E}|$, let the $k$'th edge by defined by $\mathcal{E}_k$, and let us define the incidence matrix $B \in \mathbb{R}^{|\mathcal{N}| \times |\mathcal{E}|}$ as follows
\begin{equation*}
[B]_{ik} =
\begin{cases}
	1 , & \text{ if } i \text{ is the initial node of edge } \mathcal{E}_k. \\
	-1 , & \text{ if } i \text{ is the terminal node of edge } \mathcal{E}_k. \\
0 , & \text{ otherwise}.
\end{cases}
\end{equation*}
For an undirected graph, we can define the \emph{Laplacian matrix} as $L:=BB^\top \in \mathbb{R}^{|\mathcal{N}| \times |\mathcal{N}|}$ \cite{bullo2018lectures}, and the \emph{Edge Laplacian matrix} as $L_e := B^\top B \in \mathbb{R}^{|\mathcal{E}| \times |\mathcal{E}|}$ \cite{zelazo2007agreement}. We note that $1_{|\mathcal{N}|}$ is in the kernel of $L$, and that $L_e$ is positive definite if $\mathcal{G}$ does not contain any cycles. Let us formulate the following assumption for our undirected graphs.
\begin{assumption}
\label{graph}
The undirected graph $\mathcal{G}$ is connected without any cycles.
\end{assumption}
The absence of cycles is for the conciseness of the presented results. In particular, once our results are proven for such a graph $\mathcal{G}$, then it can be extended seamlessly to general graphs by following \cite{zelazo2007agreement}, although cycles must be taken with care for closed paths \cite{sepulchre2010consensus}.

\subsection{Parametrized Path}

Consider the position $p = [p_x, p_y]^\top\in \mathbb{R}^2$ in the $2$-dimensional Euclidean space. The objective for our robots is to track a $1$-dimensional path defined as $\mathcal{P} = \{p \in \mathbb{R}^2: \phi(p) = 0 \}$, where $\phi(\cdot): \mathbb{R}^2 \rightarrow \mathbb{R}$ is the implicit equation describing the path; we consider that $\mathcal{P}$ is connected and non-empty, and that the function $\phi$ should be at least twice differentiable. If $p$ is also the robot's position, then $\phi(p) = 0$ denotes that the robot is on the path $\mathcal{P}$, and note that $\phi(p)$ is a valid error signal for a path following control law \cite{kapitanyuk2017guiding}.

Alternatively, the same path $\mathcal{P}$ can be parametrized by $p = f(x) \iff \phi(p) = 0$, where $x \in \mathbb{R}$ represents the \emph{path parameter}, and $f(\cdot): \mathbb{R} \rightarrow \mathbb{R}^2$ is a twice continuously differentiable function. In the following example, the specific case of a straight-line path is presented, as this type of trajectory will be the focus of our analysis and experiments.

\begin{example}
    The implicit equation of a straight-line path is given by $\phi(p) = (p_y - b) \sin(\alpha) - (p_x - a)\cos(\alpha)$, where $p_0 := [a, b]^\top \in \mathbb{R}^2$ represents the origin, and the slope of the line is determined by the heading angle $\alpha \in \mathbb{R}$. In the parametric form, the same path can be described as $p = f(x) = [a + \cos(\alpha)x, b + \sin(\alpha)x]^\top$. 
\end{example}

\subsection{Inverse Kinematics Guiding Vector Field}

The dynamics of the level-set error $\phi$ can be expressed as
\begin{equation}
\label{phip}
\dot{\phi}(p) = \nabla \phi^\top(p)\dot{p},
\end{equation}
where $\nabla \phi(p) \in \mathbb{R}^2$ represents the gradient of $\phi$, and note that $\nabla \phi^\top(p)$ is a Jacobian. To have a precise control of the evolution of the level-set error dynamics, we want to set $\dot{\phi}(p) = u_\phi(\phi,t)$, where $u_\phi(\phi,t): \mathbb{R} \times \mathbb{R}^+ \rightarrow \mathbb{R}$ is a \emph{virtual input} \cite{zhou2025inverse}. This desired evolution is not only to ensure convergence but also to induce a specific behavior around the path, driven by a time-varying signal $\gamma(t)$, where $\gamma(t): \mathbb{R}^+ \rightarrow \mathbb{R}$ is locally Lipschitz. The virtual input can be split as $u_\phi(\phi,t) = u_{\phi,C}(\phi) + u_{\phi,B}(t)$, where $u_{\phi, C}(\phi): \mathbb{R} \rightarrow \mathbb{R}$ is the converging input, responsible for guiding the system toward the desired path, and $u_{\phi, B} (t) : \mathbb{R}^+ \rightarrow \mathbb{R}$ is the feed-forward input governing the eventual desired behavior of a robot around the path.

A possible design for these two components is given by $u_{\phi, C}(\phi(p)) = -k_e \phi(p)$ and $u_{\phi, B}(t) = k_e\gamma(t) + \dot{\gamma}(t)$ respectively, where $k_e \in \mathbb{R}^+$ is a positive constant gain; hence, it can be deduced that
\begin{equation}
\label{dot_phi}
	\dot{\phi}(p) = u_{\phi}(\phi(p), t) = -k_e \left( \phi(p) - \gamma(t) \right)+ \dot{\gamma}(t),
\end{equation}
consequently, $\lim_{t \rightarrow \infty}{\phi}(p(t)) = {\gamma}(t)$.

Following the approach proposed in \cite{zhou2025inverse}, a vector field $f: \mathbb{R}^2 \rightarrow \mathbb{R}^2$ is designed so that the error evolves according to $\dot\phi(p) = u_\phi(\phi(p),t)$. This vector field is defined as
\begin{align} \label{gvf}
    f(\phi(p), t) &= v_T(p) + v_C(p) + v_B(t)\nonumber \\ 
    &= v_T(p) + \frac{\triangledown \phi(p)}{\|\triangledown \phi(p)\|^2}u_\phi(\phi(p),t),
\end{align}
where $v_T \in \mathbb{R}^2$ is the tangent velocity parallel to the path $\mathcal{P}$, $v_C \in \mathbb{R}^2$ is the converging velocity perpendicular to $\mathcal{P}$, and $v_B \in \mathbb{R}^2$ is the additional feed-forward term driving the desired behavior around $\mathcal{P}$ also perpendicular to $v_T$. 

In order to match the vector field to the speed of the robot, \eqref{gvf} can be accommodated as
\begin{equation}
\label{f_1}
f(\phi(p), t) = \alpha(p, t)\hat{v}_{T}(p) + \beta(p, t),
\end{equation}
where $\beta = (v_{C} + v_{B})$, and $\alpha = \sqrt{v^2 - \|\beta\|}$ is a function that will assist us to keep $\|f(\phi(p), t)\| = v$ for a given constant speed $v \in \mathbb{R}^+$. The vector field in \eqref{f_1} is a well-defined vector field within $\Omega_b := \{p \in \mathbb{R}^2 : \|\phi(p)\| \leq b, b \in \mathbb{R}^+\}$ \cite{zhou2025inverse}, where $\|v_T(p)\| > 0$ is assumed and the Jacobian $\nabla\phi(p)^\top \neq 0$ in $\Omega_b$. The vector field (\ref{f_1}) is referred to as an IK-GVF in $\Omega_b$. When $p \notin \Omega_\mathrm{b}$, we have $\|\beta\| > v$, and instead, the vector field is given by
\begin{equation}
\label{f_2}
f(\phi(p), t) = v\hat{\beta}(p,t).
\end{equation}
It can be verified that the union of the vector fields \eqref{f_1} and \eqref{f_2} results in a continuous vector field \cite{zhou2025inverse}.


\subsection{Problem formulation}

Consider a team of $N$ fixed-wing drones flying at the same constant speeds. The dynamics of the $i$'th drone are
\begin{equation}
\label{fixed-wing}
\left\{\begin{aligned}
&\dot{p}_i(t) = R(\theta_i(t))\tilde{v}\\
&\dot{\theta}_i(t) = \omega_i(t)
\end{aligned}\right., 
\quad
\forall i \in \mathcal{V},
\end{equation}
where $p_i = [p_{i,x}, p_{i.y}]^\top \in \mathbb{R}^2$ represents the position of the $i$'th drone in the plane, 
$R(\theta_i) = \left[\begin{smallmatrix}
\cos\theta_i & -\sin\theta_i \\
\sin\theta_i & \cos\theta_i \\
\end{smallmatrix}\right] = e^{E\theta_i}$, where
$E = \left[\begin{smallmatrix}
0 & -1 \\
1 & 0 \\
\end{smallmatrix}\right]$, is the heading with $\theta_i\in\mathbb{R}$, $\tilde{v} = [v, 0]^\top \in \mathbb{R}^2$ contains the constant flying speed $v$, and the heading rate $\omega_i \in \mathbb{R}$ is our actuation on the drone.

\begin{problem} \label{problem}
(\textit{Formation Flight of Fixed-Wing Drones}) Consider a team of $N$ fixed-wing drones flying at constant speeds, whose dynamics are given by \eqref{fixed-wing}, a communication graph $\mathcal{G}$ satisfying Assumption \ref{graph} where each drone $i$ is a node in $\mathcal{G}$, and a set of $N$ parametrized straight-line paths; then, design a behavior $\gamma_i(t)$ and a control law $\omega_i(t)$ for all drone $i\in\mathcal{N}$ such that:
\begin{equation}
\lim_{t\rightarrow\infty}(x_i(t) - x_j(t)) = 0, \quad \lim_{t\rightarrow\infty}\phi(p_i(t)) = 0, \quad \forall \mathcal{E}_k \in \mathcal{E}. \nonumber
\end{equation}
\end{problem}

$ $ 

\begin{figure}[t]
  \centering
  \includegraphics[width=1\hsize]{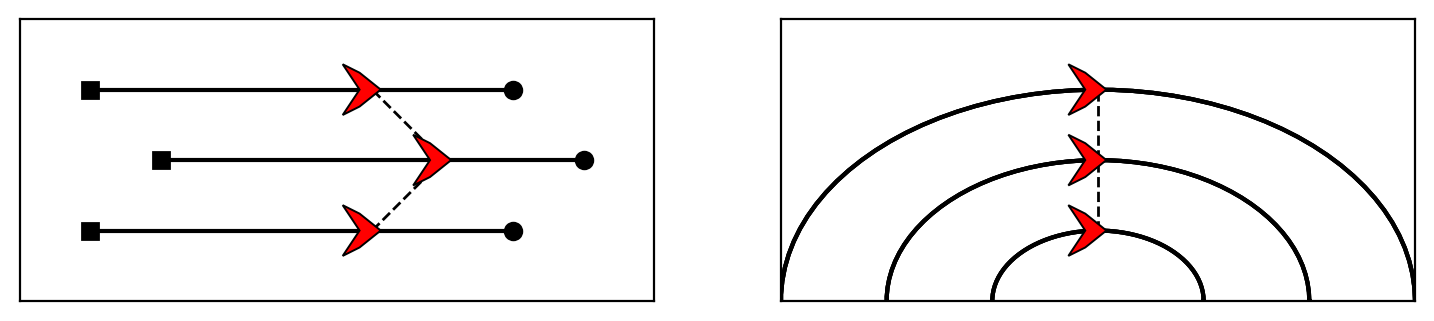}
  \caption{Two different teams of three fixed-wing drones achieve distinct formations by synchronizing their path parameters as proposed in Problem \ref{problem}. On the left, shifting the middle straight-line path's origin to the right enables a delta formation. On the right, aligning the centers of the three ellipsoidal paths allows the drones to fly parallel while orbiting the origin.}
  \label{problem_fig}
\end{figure}

\begin{remark}
The choice of straight-line paths is driven by their mathematical convenience, ease of practical implementation, and ability to effectively showcase the capabilities of this approach. Despite this choice, it will be clear that our methodology allows for a variety of different formation flight patterns with valuable practical applications, as illustrated in Figure \ref{problem_fig}. Moreover, dealing with other types of paths is straightforward since our methodology can be applied to any one-dimensional (open or closed) path in parametric form with the appropriate selection of speeds for the drones.
\end{remark}

\section{Main results} \label{sec: results}


\subsection{Design of oscillatory behavior}


We consider the following oscillatory behavior for the $i$'th drone around its path $\mathcal{P}_i$
\begin{equation} \label{gamma_eq}
    \gamma_i(t) = A_i \sin(w_\gamma t),
\end{equation}
where $A_i\in\mathbb{R}_{\geq 0}$ represents the non-constant amplitude of the oscillation that will be actuated, and $w_\gamma\in\mathbb{R}^+$ is a fixed oscillation frequency common to all the drones.

The time derivative of $\gamma_i(t)$ in \eqref{gamma_eq} is given by
\begin{equation}
\label{dotgamma}
\dot{\gamma}_i(t) = \dot{A}_i(t) \sin(w_\gamma t) + A_i(t) w_\gamma \cos(w_\gamma t).
\end{equation}

Considering $\dot p_i = f(\phi_i(p_i),t)$, with $f_i$ as in \eqref{f_1} and \eqref{f_2}, it follows from (\ref{dot_phi}) that if the drone is well aligned with the IK-GVF, then ${\phi}_i(p_i) = {\gamma}_i(t)$ in $\Omega_b$, and then $\dot{\phi}_i(p_i) = \dot{\gamma}_i(t)$, implying that $\beta_i(p_i,t) = \dot \gamma_i(t)$ in \eqref{f_1}. Consequently, the path parameter evolves as $\dot x_i(t) = \alpha(p_i,t) \hat{v}_{i,T}(p_i,t)$, and under the mild assumption that $A_i(t)$ in \eqref{dotgamma} varies slowly enough,
\begin{equation}
    \dot x_i(t) \approx \sqrt{v^2 - A_i(t)^2 w_\gamma^2 \cos^2(w_\gamma t)},
	\label{eq: xidot}
\end{equation}
where we can see that the instantaneous velocity $\dot x_i(t)$ oscillates in synchrony with $\cos(w_\gamma t)$, leading to a periodic fluctuation. Such an oscillation means that $\dot x_i$ does not provide a constant measure of progression rate along the path. Instead, the time variation of the average path parameter, denoted as the \textit{average parametric velocity} $\dot{\bar{x}}_i$ is considered. By averaging over a full oscillation period $T = \frac{2\pi}{w_\gamma}$, $\dot{\bar{x}}_i$ better captures the progression rate along the part over time, and it is defined as follows
\begin{align} \label{eq: dot_bar_x}
	\dot{\bar{x}}_i(t) &:= \frac{1}{T}\int_{0}^{T}\sqrt{v^{2} - A_i(t)^{2} w_\gamma^{2} \cos^{2}(w_\gamma t)}\mathrm{dt} \nonumber\\
&= \frac{\sqrt{v^2 - A_i(t)^2\omega_\gamma^2}}{2\pi} E\scalemath{0.98}{\left(w_\gamma x \left| \frac{A_i(t)^2 w_\gamma^2}{A_i(t)^2 w_\gamma^2 - v^2}\right.\right)},
\end{align}
where $E(\varphi|k^2) = \int_0^\varphi \sqrt{1-k^2\sin^2(\theta)} \mathrm{d}\theta$ is the incomplete elliptic integral of the second kind. 

\begin{figure}
  \centering
  \includegraphics[width=1\hsize, trim=-1 5 -1 5, clip]{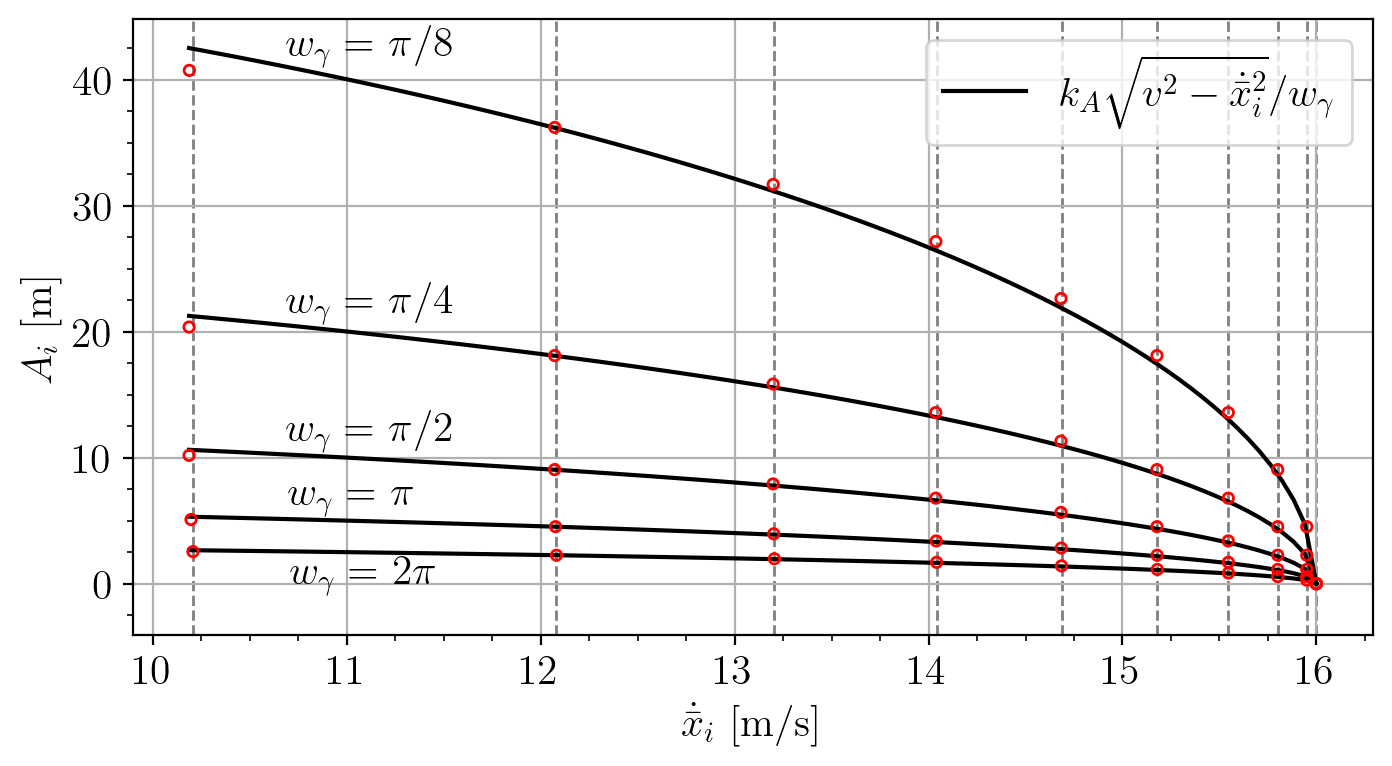}
  \caption{Fitting examples of the curves (\ref{A_i}) to the red points generated from (\ref{eq: dot_bar_x}) that show the relationship between the amplitude $A_i$ and the average parametric velocity $\dot{\bar{x}}_i$ for $v = 16$ m/s and $k_A = 1.35$ at different frequencies $w_\gamma$.}
  \label{fig_A}
\end{figure}

Leveraging the properties of elliptic functions and performing series expansions, the relationship between a time-varying amplitude $A_{i,d}(t)$ that achieves a desired time-varying average parametric velocity $\dot{\bar{x}}_{i,d}(t)$ can be approximated as
\begin{equation}
\label{A_i}
A_{i,d}(\dot{\bar{x}}_{i,d}(t)) \approx k_A \frac{\sqrt{v^2 - (\dot{\bar{x}}_{i,d}(t))^2}}{w_\gamma},
\end{equation}
where $k_A \in \mathbb{R}^+$ is a positive constant gain to be determined by a fitting procedure as in Figure \ref{fig_A}.

Observe from \eqref{eq: dot_bar_x} that negative values of $A_i(t)$ have the same effect on the average parametric velocity as positive ones. Moreover, for the oscillation to be achievable by a drone with constant speed $v$, and to ensure it remains within $\Omega_b$ when $\dot p_i = f_i$, the amplitude must satisfy the upper bound
\begin{equation}
A_i(t) \leq \frac{v}{w_\gamma}.
\end{equation}
Consequently, the average parametric velocity will be constrained as follows
\begin{equation} \label{xd_sat}
\epsilon \leq \dot{\bar{x}}_{i,d}(t) \leq v,
\end{equation}
where $\epsilon\in \mathbb{R}^+$ can be determined using the approximation in \eqref{A_i} and $A_i = v/w_\gamma$; therefore
\begin{equation} \label{epsilon}
\epsilon = \frac{\sqrt{k_A^2 - 1}}{k_A} v,
\end{equation}
and note that when $A_{i,d} = 0$, then $\dot{\bar{x}}_{i,d} = v$.

Hence, knowing how to induce an average parametric velocity, the desired average parametric velocity for the $i$'th drone to solve Problem \ref{problem} is designed as
\begin{equation}
\label{eq: dot_x_d}
\dot{\bar{x}}_{i, d}(t) = v - k_u u_i(t),
\end{equation}
where $k_u \in \mathbb{R}^+$ is a positive constant gain, and $u_i \in \mathbb{R}$ is the consensus controller to be designed considering the saturation limits in \eqref{xd_sat}. Indeed, note that if the consensus output $u_i = 0, \forall i$ means that $(x_i(t) - x_j(t)) = 0$ for all neighboring robots in $\mathcal{G}$, then $\dot{\bar{x}}_{i,d} = v$.

\subsection{Non-negative input-saturated consensus controller for oscillatory behaviors}

\begin{figure}
  \centering
  \begin{tikzpicture}

    \draw[->,black] (-1.2,0) -- (4,0) node[right] {\small $s$};
    \draw[->,black] (0,-0.2) -- (0,1.2) node[above] {\small $\tau$};

    \draw[red, thick, dashed] (-1,1) -- (3,1) node[right, yshift=4, xshift=-6] {\small $\tau_h$};
    \draw[red, thick, dashed] (-1,0) -- (3,0) node[right, yshift=4, xshift=-6] {\small $\tau_l$};

    \draw[very thick] (-0.5,0) -- (0,0);
    \draw[very thick] (0,0) -- (2,1);
    \draw[very thick] (2,1) -- (2.5,1);

    \draw[dashed] (2,0) -- (2,1) node[above right] {\small};

    \node[below] at (0.15,0) {\small $0$};
    \node[below] at (2,0) {\small $r$};

  \end{tikzpicture}
  \caption{A modified linear saturation function, as defined in \eqref{semi_line}, is shown, with the red lines representing the upper and lower bounds.}
  \label{semi}
\end{figure}
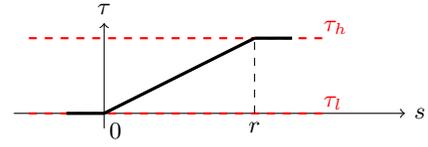

For the design of the consensus protocol $u_i$ in (\ref{eq: dot_x_d}), we consider a graph $\mathcal{G}$ under Assumption \ref{graph}, along with the constraint of the saturation in \eqref{xd_sat}. Since the output of the consensus feeds an average parametric velocity, we naturally consider as an input to $u_i$ the relative \textit{average parametric displacements}, defined for the $i$'th robot as $\bar{x}_{i} := \frac{1}{T} \int_0^T x_i(t) \mathrm{dt}$; hence,
\begin{equation}
\label{x_consensus}
u_i(t) = \mathrm{sat}_i \left[ \sum_{j \in \mathcal{N}_i}(\bar{x}_{j}(t) - \bar{x}_{i}(t)) \right],
\end{equation}
where $\mathrm{sat}_i(\cdot): \mathbb{R} \rightarrow \mathbb{R}$ is a saturation function to be designed. In particular, various non-negative saturation functions can be applied with a bounded region between zero and a positive value. Let us suggest the following \emph{linear saturation function} as illustrated in Figure \ref{semi}
\begin{equation}
\label{semi_line}
\mathrm{sat}_i(s) =
\begin{cases}
\tau_h  & \text{ if } s \geq r \\
\frac{\tau_h - \tau_l}{r}s + \tau_l & \text{ if } 0 < s < r \\
\tau_l  & \text{ if } s \leq 0
\end{cases},
\end{equation}
where $\tau_l \geq 0$ and $\tau_h > \tau_l$ define the saturation bounds, and $r>0$ is the saturation level that tunes the convergence rate together with $k_u$ in (\ref{eq: dot_x_d}).

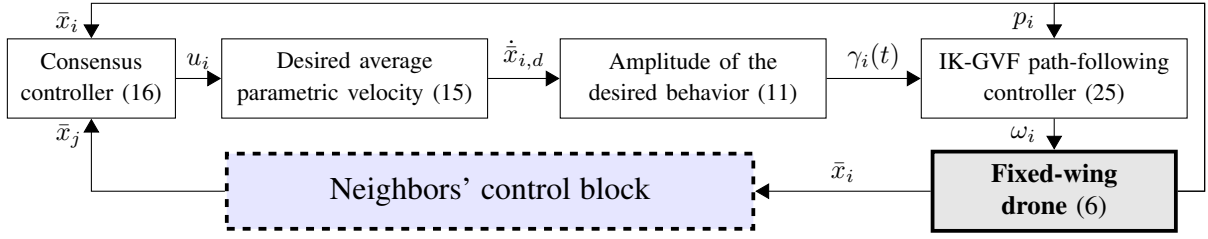
\begin{figure*}[t]
  \centering
  \begin{tikzpicture}[auto, node distance=2cm, >=triangle 60]

    \node [block_small] (cons) {{\small Consensus controller \eqref{x_consensus}}};
    \node [block, right of=cons, node distance=3.5cm] (xd) 
            {\small Desired average parametric velocity \eqref{eq: dot_x_d}};
    \node [block, right of=xd, node distance=4.5cm] (amp) 
            {\small Amplitude of the desired behavior \eqref{A_i}};
    \node [block, right of=amp, node distance=4.8cm] (path_follow) 
            {{\small IK-GVF path-following controller \eqref{heading_controller}}};
    \node [block_kin, below of=path_follow, node distance=1.5cm] (drone) 
            {{\textbf{Fixed-wing drone \eqref{fixed-wing}}}};

     \node [block_nei, left of=drone, node distance=7.5cm] (nei) 
     {{\large Neighbors' control block}};

    \draw [->] (cons) -- node[name=u] {$u_i$} (xd);
    \draw [->] (xd) -- node[name=u] {$\dot{\bar{x}}_{i,d}$} (amp);
    \draw [->] (amp) -- node[name=gamma] {$\gamma_i(t)$} (path_follow);
    \draw [->] (path_follow) -- node[name=omega, xshift=-0.7cm] {$\omega_i$} (drone);
    \draw [->] (drone.west)  |-  node [pos=0.75, yshift=0.5cm] {$\bar{x}_i$} (nei);
    \draw [->] (nei)  -|  node [pos=0.88] {$\bar{x}_j$} (cons);

    \node [point, name=output, above of=path_follow, node distance=1cm, xshift=2cm] {};
    \draw [->] (drone.east) -| (output) -| node [pos=0.5, below, xshift=-0.4cm] {$p_i$} (path_follow);
    \draw [->] (drone.east) -| (output) -| node [pos=0.5, below, xshift=-0.32cm] {$\bar{x}_i$} (cons);
        
  \end{tikzpicture}
  \caption{Control block diagram for the $i$'th fixed-wing drone.}
  \label{fig_control}
\end{figure*}

Let us first show the effectiveness of the consensus protocol \eqref{x_consensus}-\eqref{semi_line}. Without loss of generality, let us consider a system of $N$ single-integrator agents, i.e.,
\begin{equation}
\label{mas}
\dot{\bar x}_i(t) = u_i(t), i \in \mathcal{V}.
\end{equation}

\begin{theorem}
\label{theorem1}
Consider a multi-agent system \eqref{mas}, an undirected graph $\mathcal{G}$ satisfying Assumption \ref{graph}, and the consensus protocol (\ref{x_consensus})-(\ref{semi_line}) with $\tau_l = 0$. Then $\lim_{t\rightarrow\infty}(x_i(t) - x_j(t)) = 0, \forall \mathcal{E}_k \in \mathcal{E}$ and $\lim_{t\rightarrow\infty} u_i(t) = 0, \forall i\in\mathcal{V}$ for any $x_i(0) \in\mathbb{R}, \forall i\in\mathcal{V}$.
\end{theorem}

\begin{proof}
We will exploit the edge Laplacian matrix $L_e$ to prove our statement; therefore, we will focus on the relative states between neighboring agents; namely
\begin{equation*}
z_k := \bar x_i - \bar x_j, \quad \forall \mathcal{E}_k \in \mathcal{E},
\end{equation*}
	where the compact form of the relative states can be found as $z = B^\top \bar x$ with $z\in \mathbb{R}^{|\mathcal{E}|}$. By stacking $\bar x = [\bar x_1, \cdots, \bar x_{|\mathcal{N}|}] \in \mathbb{R}^{|\mathcal{N}|}$, $u = [u_1, \cdots, u_{|\mathcal{N}|}] \in \mathbb{R}^{|\mathcal{N}|}$, and defining the element-wise saturation function $\mathrm{sat}(\cdot) = [\mathrm{sat}_1(\cdot), \mathrm{sat}_2(\cdot), \cdots, \mathrm{sat}_{|\mathcal{N}|}(\cdot)] \in \mathbb{R}^{|\mathcal{N}|}$; the compact form of (\ref{mas}) can be written as
\begin{equation}
\label{dot_x2}
\dot{\bar x} = \mathrm{sat}(-L \bar x),
\end{equation}
where $L$ is the Laplacian matrix.

Now, we are going to study whether the origin of the relative state dynamics $\dot{z} = B^\top u$ is asymptotically stable, i.e., whether $\lim_{t\rightarrow\infty} z_k(t) \rightarrow 0$ for all $\mathcal{E}_k \in \mathcal{E}$. In this case, note that (\ref{x_consensus}) can be written as
\begin{equation}
	u_i = \mathrm{sat}_i \left( -\sum_{k = 1}^{|\mathcal{E}|} [B]_{ik} z_k \right),
\end{equation}
whose compact form is
\begin{equation}
\label{u_edge}
u = \mathrm{sat}(-B z),
\end{equation}
so that we arrive at the following autonomous system
\begin{equation}
\label{system_edge}
\dot{z} = B^\top \mathrm{sat}(-B z).
\end{equation}
Let $\eta := -B z$, and $\eta := [\eta_1, \dots, \eta_{|\mathcal{N}|}] \in \mathbb{R}^{|\mathcal{N}|}$; thus,
\begin{equation}
\label{dot_eta}
\dot{\eta} = -L \mathrm{sat}(\eta).
\end{equation}
By defining a \emph{classic} saturation function
\begin{equation*}
\overline{\mathrm{sat}}_i(\bar{s}) 
	= \mathrm{sat}_i\left(\bar s + \frac{r}{2}\right) - \frac{\tau_h}{2},
\end{equation*}
i.e, $s = \bar s + \frac{r}{2}$ so that $\overline{\mathrm{sat}}_i(\bar{s})$ is an odd function since $\tau_l = 0$. Then, the modified linear saturation function \eqref{semi_line} can always be a transformation of this odd saturation function as
\begin{equation*}
\mathrm{sat}_i(s) = \overline{\mathrm{sat}}_i\left(s - \frac{r}{2}\right) + \frac{\tau_h}{2}.
\end{equation*}
Let us choose the following Lyapunov function candidate as
\begin{equation}
\label{V}
V = \sum_{i = 1}^{N} \int_{0}^{\bar{\eta}_i} \overline{\mathrm{sat}}_i(\bar{s}) \mathrm{d}\bar{s},
\end{equation}
	where $\bar{\eta}_i = \eta_i - \frac{r}{2}$, and it can be verified that $V \geq 0$ by the properties of the odd saturation function. By defining $\bar{\eta} = [\bar{\eta}_1, \dots, \bar{\eta}_{\mathcal{N}}] \in \mathbb{R}^{|\mathcal{N}|}$ and taking the time derivative of \eqref{V}, one has
\begin{equation*}
\label{dot_V}
\begin{aligned}
\dot{V} &= \overline{\mathrm{sat}}^\top(\bar{\eta}) \dot{\bar{\eta}} \\
	&= \mathrm{sat}^\top (\eta) \dot{\eta} - \frac{\tau_h}{2}1_{|\mathcal{N}|}^\top \dot{\eta} \\
	&= -\mathrm{sat}^\top (\eta)\,L\,\mathrm{sat} (\eta) + \frac{\tau_h}{2}1_{|\mathcal{N}|}^\top(\eta)\,L\,\mathrm{sat} (\eta)  \\
&= -\mathrm{sat}^\top (\eta)\,L\,\mathrm{sat} (\eta) \leq 0,
\end{aligned}
\end{equation*}
	and from Lasalle's invariance principle, all the trajectories of \eqref{dot_eta} will converge to the invariant set $\{\eta: \mathrm{sat}(\eta) = c1_N\}$, where $c$ is a constant depending on the initial condition $\bar\eta(0)$. However, note that only $\eta = 0$ can make $Bz = 0$, i.e., $c$ must be zero; this is because the edge Laplacian matrix $L_e = B^\top B$ is positive definite under Assumption \ref{graph}. Consequently, the origin of the system \eqref{system_edge} is asymptotically stable, and therefore $(\bar x_j(t) - \bar x_i(t)) \rightarrow 0$ as $t \to \infty$. Finally, we can conclude that $u_i(t) \to 0$ as $t\to\infty$ since $\mathrm{sat}(0) = 0$ because $\tau_l = 0$.
\end{proof}

\begin{remark}
\label{rem_ustar}
The parameters of the linear saturation function in \eqref{semi_line} while used in \eqref{eq: dot_x_d} must fit into the saturation limits in \eqref{xd_sat}; thus we choose $\tau_l = 0$, and $\tau_h = (v - \epsilon) / k_u$. Consequently, the required amplitude is $0$ when $u_i = \tau_l$, and reach its maximum of $v/w_\gamma$ when $u_i = \tau_h$.

In this particular case, since $\tau_l = 0$, if there exists an agent $q\in\mathcal{V}$ such that
$
\bar x_q(0) = \bar x_{\max}(0),
$
then it can be verified that $\dot{\bar x}_q(t) = 0, \forall t$; therefore, all the states converge to
\begin{equation*}
\label{x_star}
\bar x^* = \bar x_q(0) = \bar x_{\max}(0),
\end{equation*}
	where again all the control inputs $u_i(t)$ converge to zero as stated in Theorem \ref{theorem1}.
\end{remark}

\subsection{Following the desired oscillatory behavior}

So far we have designed the consensus controller \eqref{x_consensus} for \eqref{eq: dot_x_d} that commands the desired average parametric velocity for each drone, i.e., it directly sets the amplitude for the oscillation $\gamma_i(t)$ as in (\ref{gamma_eq}), and together with the constant speed $v$, it defines the vector field $f_i$ as described in \eqref{f_1}-\eqref{f_2}.

In this section, we design a heading controller to align the fixed-wing drones with the resulting $f_i$ to solve Problem \ref{problem}. For clarity, the block diagram for the control system of the $i$'th drone is shown in Figure \ref{fig_control}. Following \cite[Theorem 2]{zhou2025inverse}, the heading controller can be developed for the $i$'th drone as
\begin{equation}
\label{heading_controller}
\begin{aligned}
\omega_i(t) =& k_n \xi_i(\phi_i(p_i),t)\dot{p}_i(t) + \dot{\theta}_{i, d}(t)
\\
=& \xi_i(\phi_i(p_i),t)\left[ k_n \dot{p}_i(t) - \dot{f}_i(\phi_i(p_i), t) \right],
\end{aligned}
\end{equation}
where $\xi_i = \frac{f_i^\top E}{v^2}$, $\dot{\theta}_{i, d} = \frac{f_i^\top E^\top}{v^2}\dot{f}_i = - \xi_i \dot{f}_i$ captures the variation of the designed vector field, and $k_n \in \mathbb{R}^+$ is a positive constant gain to tune the convergence rate. For the time variation of $\dot f_i$, if $\|\beta_i\| > v$,
\begin{equation*}
\dot{f}_i = \left( \frac{I_2}{\|\beta_i\|} - \frac{\beta_i\beta_i^\top}{\|\beta_i\|^3} \right) \dot{\beta}_i v,
\end{equation*}
otherwise,
\begin{equation*}
\dot{f}_i = \dot{\alpha}_i \hat{v}_{i,T} + \alpha_i \left( \frac{I_2}{\|v_{i,T}\|} - \frac{v_{i,T}v_{i,T}^\top}{\|v_{i,T}\|^3} \right) \dot{v}_{i,T} + \dot{\beta}_{i},
\end{equation*}
where
$\dot{\beta}_i = \dot{\zeta}_i (u_{i,\phi} + \dot{\gamma_i}(t)) + \zeta_i (\dot{u}_{i,\phi} + \ddot{\gamma_i}(t))$,
with 
$\zeta_i = \frac{\triangledown\phi_i}{\|\triangledown\phi_i\|^2}$, 
and based on the Hessian matrix $H_{i,\phi}$, one has
\begin{equation*} 
\dot{\zeta}_i = \frac{H_{i,\phi}^\top \dot{p}_i - \zeta_i (\dot{p}_i^\top H_{i,\phi} \triangledown\phi_i + \triangledown\phi_i^\top H_{i,\phi}^\top f_i)}{\|\triangledown\phi_i\|^2},
\end{equation*}
and we note that for the straight line, its Hessian is the zero matrix, i.e., no curvature. Also note that we capture the variation of the amplitude of $\gamma_i$ driven by \eqref{eq: dot_x_d} in $\dot\beta_i$ since it needs $\dot\gamma_i$ and $\ddot\gamma_i$ for its computation.

Considering \eqref{heading_controller} as the path-following controller for the vector field $f_i$ defined by \eqref{f_1} and \eqref{f_2}, \cite[Theorem 2]{zhou2025inverse} already establishes that $\lim_{t\rightarrow\infty}\dot{p}_i(t) = f_i(\phi(p_i),t)$, $\forall i \in \mathcal{V}$, ensuring that $\lim_{t\rightarrow\infty}\phi(p_i(t)) = 0$, $\forall i \in \mathcal{V}$ due to the design of $f_i$. Note that this holds even when $\dot{A}_i(t) \neq 0$, as the path-following controller in \eqref{heading_controller} inherently accounts for the time variation of $\gamma(t)$. Hence, since the oscillatory behavior $\gamma_i(t)$ in \eqref{gamma_eq} is designed according to the amplitude $A_i(t)$ required in \eqref{A_i} by the desired average parametric velocity $\dot{\bar{x}}_i$ in \eqref{eq: dot_x_d}, it follows that $\lim_{t\rightarrow\infty}\dot{\bar{x}}_i(t) = \dot{\bar{x}}_{i,d}(t)$, $\forall i\in\mathcal{V}$. Furthermore, from Theorem \ref{theorem1} and Remark \ref{rem_ustar}, the consensus law \eqref{x_consensus}, which regulates the average parametric velocity $\dot{\bar{x}}_{i,d}(t)$ as in \eqref{eq: dot_x_d}, ensures that $\lim_{t\rightarrow\infty}(\bar x_i(t) - \bar x_j(t)) = 0, \forall \mathcal{E}_k \in \mathcal{E}$. Nonetheless, such a asymptotic value corresponds to $u(t) = 0$, i.e., $A_i = 0, \forall i$, and then $\dot{\bar x}_i = \dot x_i$; consequently, we can conclude that $\lim_{t\rightarrow\infty}(x_i(t) - x_j(t)) = 0, \forall \mathcal{E}_k \in \mathcal{E}$, solving in this way Problem \ref{problem}. Note that if the drones follow a generic path $\mathcal{P}$, then, in order to achieve consensus for their path parameters, the drones' speeds must be designed accordingly to have a feasible formation flight. We summarize this analysis with our main result.

\begin{theorem}
	Consider a team of $N$ fixed-wing drones with dynamics (\ref{fixed-wing}) and equal constant speed $v$, whose inter-communication is given by an undirected graph $\mathcal{G}$ with Assumption \ref{graph}. If the drones actuate over their headings with (\ref{heading_controller}) whose IK-GVF $f_i$ is driven by $\gamma_i$ as defined in (\ref{gamma_eq}), and its amplitude $A_i$ is given by (\ref{A_i}) through \eqref{eq: dot_x_d}-\eqref{x_consensus}, then, for a desired straight-line path $\mathcal{P}$, Problem \ref{problem} is solved.
\end{theorem}

\begin{remark}
	The omission of saturation on the heading angular speed $\omega_i$ in the controller design is justified by assuming that all control inputs remain within the nominal limits of the fixed-wing drones. As we will see during the real experiments, this is a realistic assumption for the demanded (\refeq{heading_controller}). This assumption is necessary to ensure the controller performs as intended, allowing the drones to follow the desired oscillatory behavior accurately. If the heading dynamics were to saturate, the drones might not be able to track the required oscillations properly, potentially leading to divergence from the desired paths.
\end{remark}

\section{Flight experiments}
\label{sec: exp}

\subsection{Experimental Setup}

\begin{figure}[t]
  \centering
  \includegraphics[width=0.8\hsize]{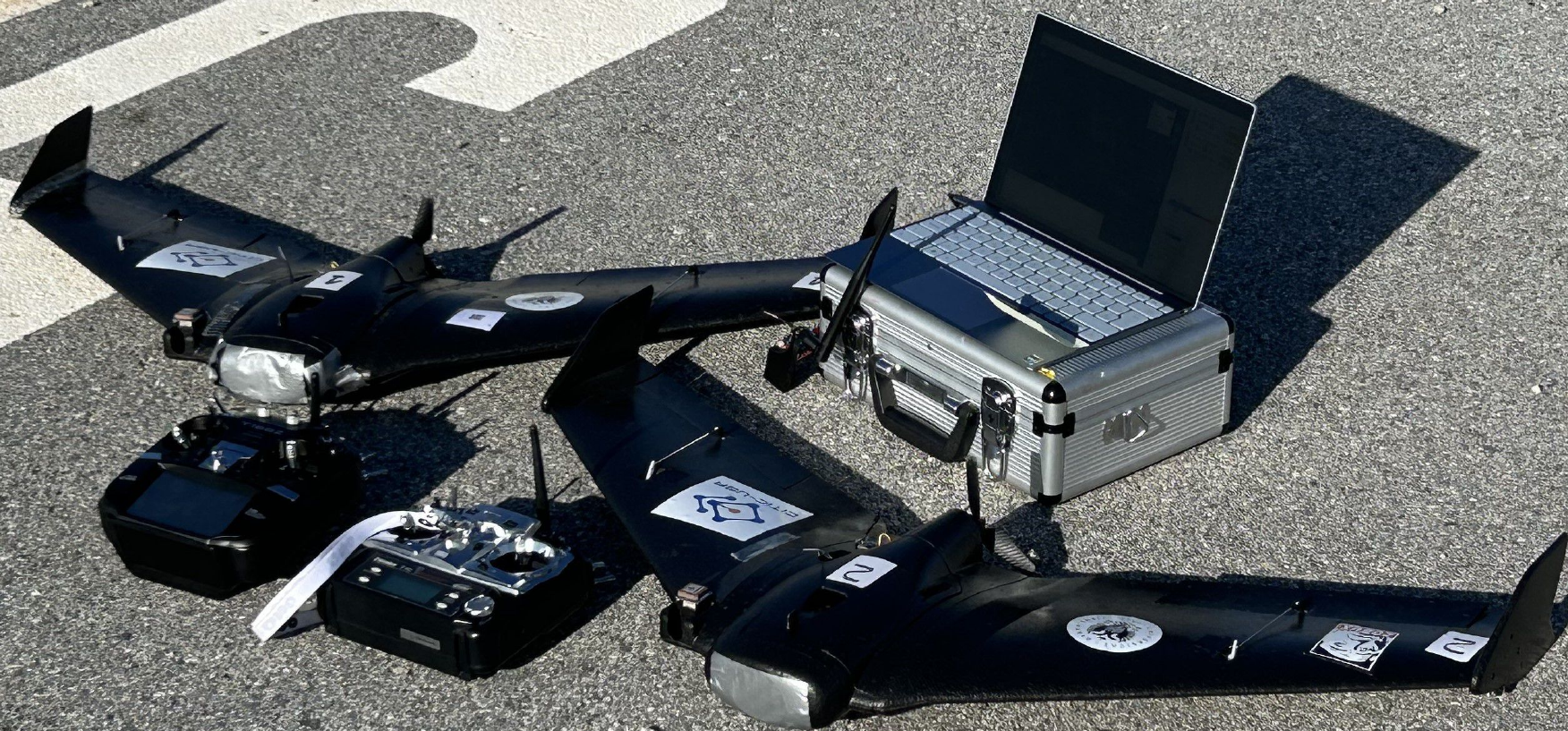}
  \caption{Fixed-wing drones deployed in the flight experiments.}
  \label{fig_UAV}
\end{figure}

The two fixed-wing drones deployed are Sonicmodell AR Wing aircraft made of EPP foam, with a $0.9$m wingspan, as illustrated in Figure \ref{fig_UAV}. Each is equipped with a $2200$kV motor, a pair of $9$g servos, a Ublox GPS module, a $3$S $2200$mAh LiPo battery, and a Futaba transmitter for secure manual override. Telemetry is handled via an Xbee Pro S1 module, while an Apogee board running Paparazzi firmware serves as the flight controller. Weighing $650$g, the drone can fly for up to 25 minutes at $50$m above ground (approximately $720$m above mean sea level) with a cruising speed of around $16$m/s. The ground control system consists of a laptop linked to another Xbee Pro S1 with a dipole antenna, and all are managed through Paparazzi software \cite{paparazzi2024}. Test flights were conducted at the \emph{Ciudad de la Alhambra} club in Granada, Spain.

\begin{figure}[t]
  \centering
  \includegraphics[width=1\hsize, trim=0 5 0 5, clip]{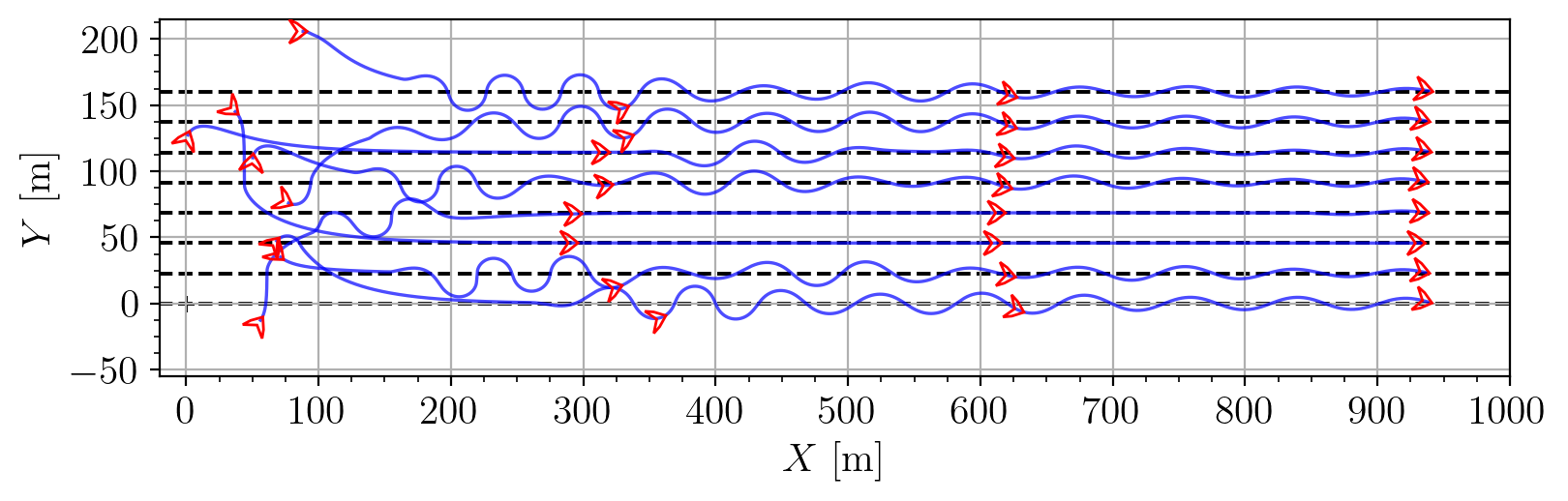}
  \caption{A team of $N=8$ fixed-wing drones moving at the same constant speed $v=8$m/s and connected in a spanning tree graph with edges $\mathcal{E} = ((1,2), (2,3), (3,4), (3,5), (4,6), (5,7), (6,8))$, is commanded to follow distinct straight-line paths while synchronized at common path parameter. For the desired average parametric velocity $\dot{\bar{x}}_{i,d}$ in \eqref{eq: dot_x_d}, $k_u=0.16$ is used, and for the corresponding  $A_{i,d}$ in \eqref{A_i}, $k_A = 1.35$ as estimated in Figure \ref{fig_A}. The common frequency for oscillatory behavior is $w_\gamma = 0.6$rad/s.}
  \label{fig_sim}
\end{figure}

\begin{figure*}[ht]
  \centering
  \includegraphics[width=2\columnwidth]{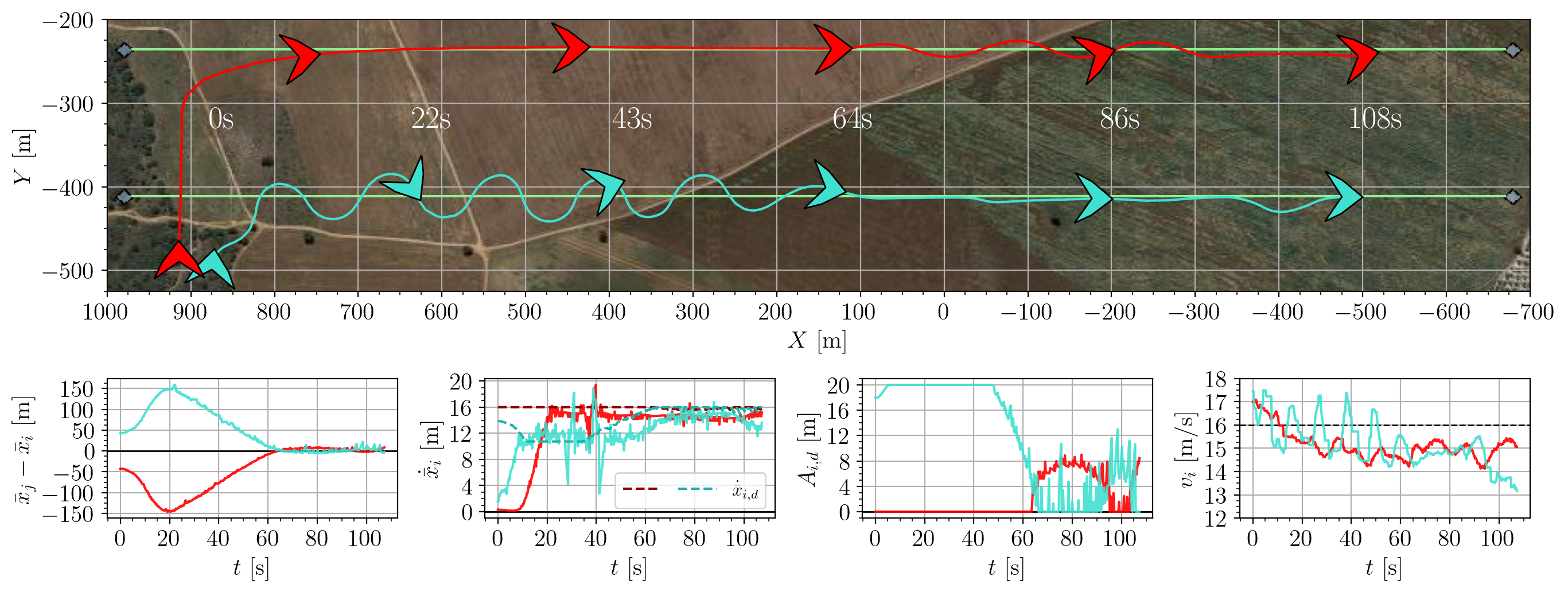}
  \caption{Telemetry data from an experiment where the two fixed-wing drones shown in Figure \ref{fig_UAV}, initially flying in formation, are commanded to follow two distinct straight-line paths with the same slope but separated by $175$m vertically. The control block, shown in Figure \ref{fig_control}, is implemented with parameters $k_u = 0.06$, $k_A = 1.35$, and $w_\gamma = 0.6$rad/s as the frequency for the oscillatory behavior. The top plot depicts the X-Y position throughout the mission, highlighting six equidistance time instants. Below, from left to right, the consensus error $(\bar{x}_i - \bar{x}_j)$ for each drone, the actual average parametric velocity $\dot{\bar x}_i$ (solid line) compared with the desired one $\dot{\bar x}_{i,d}$ (dashed line), the desired amplitude $A_{i,d}$ defining the oscillatory behavior $\gamma_i(t)$, and the actual ground speed, where the nominal constant speed of $v=16$m/s is indicated by a black dashed line.}
  \label{fig_exp}
\end{figure*}

The desired paths in the flight experiment are selected as parameterized straight-line paths. Before carrying out the flight test, it is necessary to find the suitable gain $k_A$ in (\ref{A_i}). Considering $v = 16$m/s as the nominal speed and selecting $w_\gamma = 0.6$rad/s as the oscillation frequency, the corresponding amplitude curve can be interpolated based on the average parametric velocity, as illustrated in Figure \ref{fig_A}.
Prior to the experimental scenario, the simulation in Figure \ref{fig_sim} numerically verifies the performance of these estimations with $8$ drones\footnote{At the moment of the submission, the code can be found at \url{https://github.com/Swarm-Systems-Lab/gvf_ik/tree/consensus}}.

\subsection{Experimental Analysis}

The experimental results presented in Figure \ref{fig_exp} illustrate a scenario where the two fixed-wing drones, initially flying in formation, are commanded to follow two distinct straight-line paths with the same slope but separated by $175$m. Using the control block illustrated in Figure \ref{fig_control}, the faster drone injects a significant oscillation to wait for the slower drone until they synchronize at a common path parameter to preserve their original formation. Once synchronized, both drones advance at similar forward speeds, despite variations in their ground speeds due to crosswind disturbances of approximately $3$-$4$m/s and the altitude controller, which requires different throttle adjustments when the altitude deviates above or below the designated flight altitude.

Although both drones assume a constant speed of $v=16$m/s for $A_{i,d}$ in \eqref{A_i}, the data represented in Figure \ref{fig_exp} demonstrate that the commanded $A_{i,d}$, with the estimated $k_A = 1.35$ from Figure \ref{fig_A}, successfully achieves the desired average parametric velocity $\dot{\bar{x}}_{i,d}$. This confirms our hypothesis that our algorithm does not enforce constant speeds but they are free parameters that we do not need to actuate for the coordinated formation flight. The observed delay between the desired average parametric velocity and the actual one (see second plot from the left in Figure \ref{fig_exp}) arises from working with averages, which inherently introduce a lag equal to the oscillation period. Since the control strategy in \eqref{x_consensus} is based on average values rather than instantaneous displacements, it effectively functions as a pure integral controller. This characteristic explains both the extended convergence time and the presence of steady-state oscillations. 

Finally, note that the maximum desired amplitude is software-limited to $20$m which is lower than $A_{i,d}(\epsilon) = 26.7$m, where we have considered \eqref{A_i} and \eqref{epsilon} for such a computation. This is because experimental results indicate that at a frequency of $w_\gamma = 0.6$rad/s, oscillations with amplitudes exceeding $A=20$m are challenging for our drones to follow.

\section{Conclusions}


This paper presents a novel approach to distributed formation flight for fixed-wing drones. By designing a fixed-frequency oscillatory behavior, with its amplitude modulated through a non-negative input-saturated consensus strategy, and integrating a heading controller to align each drone with the desired oscillatory behavior while converging to the desired path, the proposed method ensures coordinated formation flight while maintaining synchronized path following, without requiring actuation on the drones' speed. Experimental validation with real fixed-wing drones confirms the effectiveness of the approach. Future research could explore several avenues for improvement. Two potential directions include incorporating a proportional term into the consensus controller to mitigate delays inherent in the actual implementation, thereby enhancing the overall stability and responsiveness of the system, and analyzing the effects of saturation on the heading angular speed. Additional avenues involve  compensation strategies for velocity mismatches between agents to reduce steady-state errors, and exploring how minimal actuation based on relative velocity differences can enable a broader class of formation behaviors.



\bibliographystyle{IEEEtran}
\bibliography{mybibfile}

\end{document}